\documentclass[submission]{article} 
\usepackage{aaai2026}  
\usepackage{times}  
\usepackage{helvet}  
\usepackage{courier}  
\usepackage[hyphens]{url}  
\usepackage{graphicx} 
\usepackage{caption} 
\usepackage[table]{xcolor}
\usepackage{natbib}  
\usepackage{caption} 
\usepackage{algorithm}
\usepackage{algorithmic}

\usepackage{newfloat}
\usepackage{booktabs}
\usepackage{listings}
\usepackage[most]{tcolorbox}
\definecolor{01}{RGB}{72,192,170}
\definecolor{02}{RGB}{69,105,144}
\definecolor{03}{RGB}{239,118,122}
\definecolor{p1}{RGB}{158,129,186}
\definecolor{p2}{RGB}{216,207,227}
\definecolor{p4}{RGB}{130,176,210}
\definecolor{p3}{RGB}{40,120,181}
\usepackage{amssymb}
\usepackage{bm}
\DeclareCaptionStyle{ruled}{labelfont=normalfont,labelsep=colon,strut=off} 
\floatstyle{ruled}
\newfloat{listing}{tb}{lst}{}
\floatname{listing}{Listing}
\title{MEJO: MLLM-Engaged Surgical Triplet Recognition via Inter- and Intra-Task Joint Optimization}
\author {
    Yiyi Zhang\textsuperscript{\rm 1},
    Yuchen Yuan\textsuperscript{\rm 1},
    Ying Zheng\textsuperscript{\rm 2},
    Jialun Pei\textsuperscript{\rm 1 }, 
    Jinpeng Li\textsuperscript{\rm 1 \thanks{Corresponding Author}},
    Zheng Li\textsuperscript{\rm 3},
    Pheng-Ann Heng\textsuperscript{\rm 1}
}
\affiliations {
    \textsuperscript{\rm 1}Department of Computer Science and Engineering, The Chinese University of Hong Kong\\
    \textsuperscript{\rm 2}Department of Electrical and Electronic Engineering, The Hong Kong Polytechnic University\\
    \textsuperscript{\rm 3}Department of Surgery, The Chinese University of Hong Kong\\
    yyzhang24@cse.cuhk.edu.hk
}

\usepackage{bibentry}

\begin{document}

\maketitle

\begin{abstract}
Surgical triplet recognition, which involves identifying instrument, verb, target, and their combinations, is a complex surgical scene understanding challenge plagued by long-tailed data distribution. The mainstream multi-task learning paradigm benefiting from cross-task collaborative promotion has shown promising performance in identifying triples, but two key challenges remain: 1) inter-task optimization conflicts caused by entangling task-generic and task-specific representations; 2) intra-task optimization conflicts due to class-imbalanced training data. To overcome these difficulties, we propose the MLLM-Engaged Joint Optimization (MEJO) framework that empowers both inter- and intra-task optimization for surgical triplet recognition. For inter-task optimization, we introduce the Shared-Specific-Disentangled (S$^2$D) learning scheme that decomposes representations into task-shared and task-specific components. 
To enhance task-shared representations, we construct a Multimodal Large Language Model (MLLM) powered probabilistic prompt pool to dynamically augment visual features with expert-level semantic cues. 
Additionally, comprehensive task-specific cues are modeled via distinct task prompts covering the temporal-spatial dimensions, effectively mitigating inter-task ambiguities. To tackle intra-task optimization conflicts, we develop a Coordinated Gradient Learning (CGL) strategy, which dissects and rebalances the positive-negative gradients originating from head and tail classes for more coordinated learning behaviors. Extensive experiments on the CholecT45 and CholecT50 datasets demonstrate the superiority of our proposed framework, validating its effectiveness in handling optimization conflicts.
\end{abstract}

\section{Introduction}
\begin{figure}[t]
\includegraphics[width=8cm]{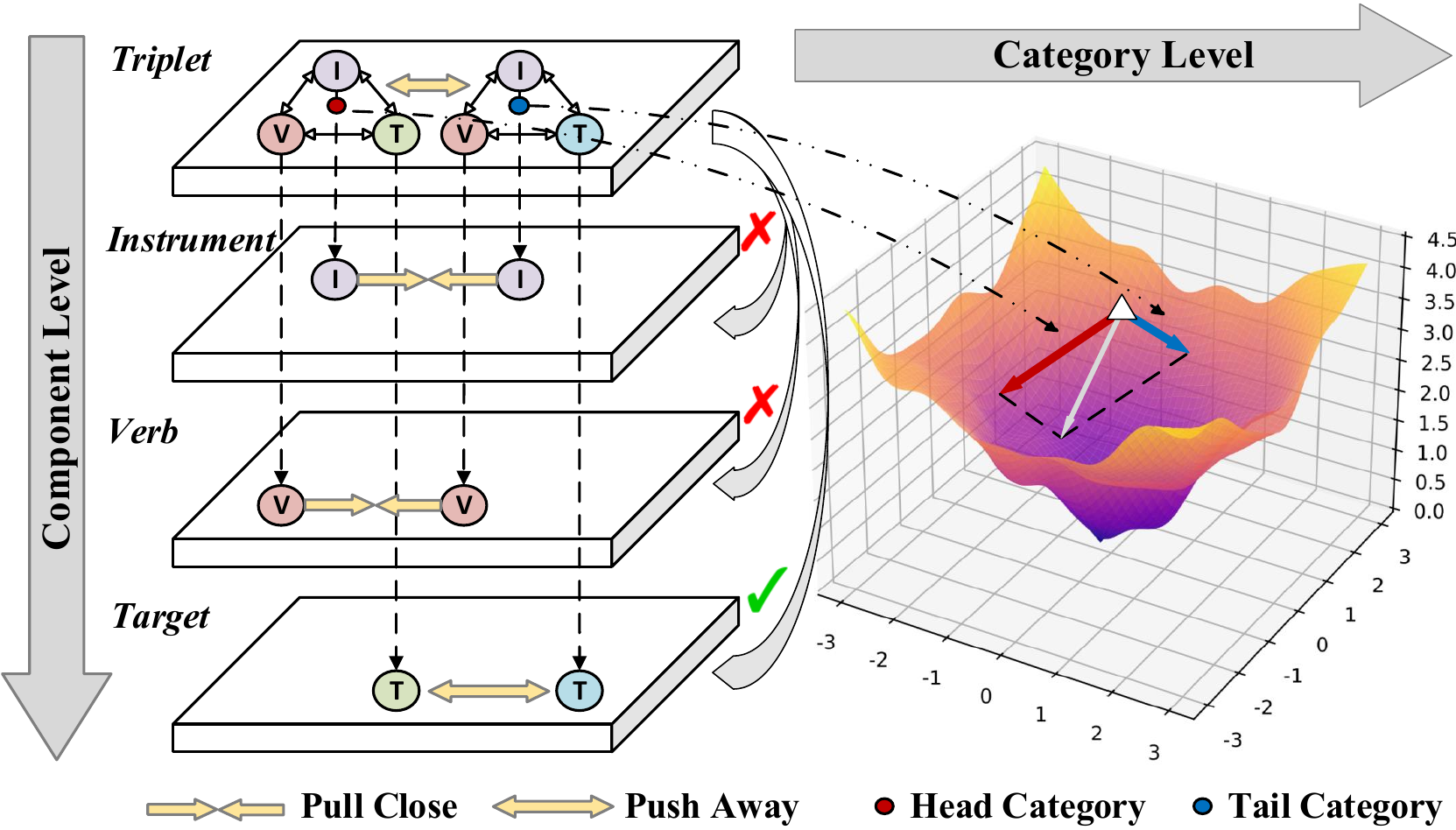}
\caption{STR faces conflicts in 1) inter-task optimization at the component level between tasks (circles with identical color represent the same category; a red cross indicates inconsistent optimization objectives); 2) intra-task conflicts at the category level between head-tail classes, where gradients from head classes dominate optimization direction.}
\label{img:teaser}
\end{figure}
In minimally invasive surgery, particularly laparoscopic cholecystectomy, precise AI-assisted decision-making is crucial for improving surgical efficiency. To alleviate the heavy burden placed on surgeons by the limited field of view and lack of tactile feedback, Surgical Triplet Recognition (STR) \cite{nwoye2020recognition} has emerged as a key technique for understanding fine-grained surgical video workflows. STR aims to identify the essential components of surgical procedures structured as $<$instrument, verb, target$>$ triplets from each video frame \cite{nwoye2023cholectriplet2021}. Mainstream methods \cite{ nwoye2022rendezvous, gui2024tail} typically disentangle STR into four distinct tasks, treating instrument, verb, and target recognition as auxiliary tasks that facilitate triplet recognition. This paradigm is well-motivated, as the auxiliary tasks provide complementary supervision that is crucial for accurately modeling the highly interrelated triplet components. Notably, STR is fundamentally \textit{instrument-centric} \cite{nwoye2022rendezvous}. This principle dictates that anatomical structures, such as the liver during laparoscopic cholecystectomy, are recognized as targets only during active instruments act upon them, regardless of their visibility. Verbs are similarly defined by active instrument-target interaction. Therefore, prevailing methodologies~\cite{sharma2023surgical,nwoye2020recognition} predominantly extract the class activation map (CAM) or bounding box of instruments to provide weakly-supervised localization cues, which are then fused into other task-specific branches to enhance the overall performance.

By utilizing auxiliary tasks to acquire supplementary supervision and leveraging instrument location cues to boost performance, existing multi-task solutions ~\cite{sharma2023rendezvous,gui2024tail,gui2024mt4mtl,peiinstrument} have shown promising results on the STR task. However, three critical problems have not been fully addressed:
\textbf{1) Inter-task optimization conflicts}. The component-decomposed tasks in STR may lead to conflicting optimization objectives during training. As illustrated in Fig. \ref{img:teaser}, one pair of frames from two different triplet categories (e.g., labeled as $<$grasper, grasp, gallbladder$>$ and $<$grasper, grasp, liver$>$)  may appear similar and should be pulled closer for instrument and verb recognition tasks. However, they should be pushed apart for target and triplet recognition tasks. These conflicting gradients of opposing objectives between auxiliary and primary tasks make joint optimization more challenging.
\textbf{2) Intra-task optimization conflicts}, which originate from gradient conflicts between head and tail classes caused by severe long-tailed data distribution. For instance, in the CholecT45 dataset \cite{nwoye2022data}, the most frequent triplet category includes more than 40,000 samples, while the least frequent class contains only 8 samples. Gradients derived from head classes dominate the overall optimization direction, leading to asynchronous learning dynamics across different classes. In this case, head classes converge rapidly and tend to overfit, while tail classes remain underfitted, eventually compromising the STR performance. 
\textbf{3) Lack of expert knowledge integration}. Conventional STR methods utilize coarse localization cues from instruments to enhance overall task learning, but largely overlook fine-grained visual details, such as the instrument’s shape, which are essential for accurate and reliable recognition. 
Leveraging insights from \cite{yue2023surgicalpart}, domain expertise regarding the instrument can be conceptualized as a composition of decoupled structures (e.g., tip, wrist, and shaft). These structures encapsulate detailed visual characteristics but require labor-intensive annotations. Fortunately, recent advancements in Multimodal Large Language Models \cite{hurst2024gpt} have facilitated the efficient extraction of fine-grained knowledge given target images.

To tackle the outlined problems, we propose the MLLM-Engaged Joint Optimization (MEJO) framework to mitigate inter-task and intra-task conflicts while effectively integrating multimodal expert knowledge. To address inter-task conflicts, the Shared-Specific-Disentangled (S$^2$D) representation learning scheme is proposed to decouple feature learning into two complementary stages:
1) Learning the informative task-shared representation that captures commonality across all tasks. 
Crucially, to instill task-shared features with multimodal expert knowledge, we leverage MLLM to construct instrument-anchored probabilistic prompt pools and dynamically select textual semantic prompts, providing more fine-grained high-level semantic guidance. 
2) Modeling task-specific representations tailored to minimize ambiguities from conflicting task objectives. We design unique temporal-spatial task prompts that facilitate the extraction of discriminative features targeted to individual tasks.
To address intra-task conflicts, we introduce a Coordinated Gradient Learning (CGL) strategy that tackles the optimization challenges arising from the severe long-tailed distribution within the triplet task.
By decomposing and rebalancing positive-negative gradients from head and tail classes, this approach fosters a more coordinated learning behavior between head-tail categories, mitigating the asynchronous convergence that has long hindered the fair learning at the category level in STR. The main contributions of this paper are as follows:

\begin{itemize}
\item We propose a MEJO framework for STR. MEJO identifies two key challenges in STR and tackles the inter-task conflicts with S$^2$D representation learning scheme and mitigates the intra-task conflicts with the CGL strategy. 
\item S$^2$D representation learning scheme is introduced to separately model task-shared and task-specific representations. Task-shared representations are dynamically enriched by MLLM-driven knowledge. The task-specific representations are learned by temporal-spatial prompts.
\item CGL is adopted to mitigate the intra-task optimization conflicts by achieving a synchronous learning behavior across head-tail triplet classes, resulting in improved tail performance and comparable head performance.
\item Extensive experiments demonstrate the effectiveness of MEJO, which achieves state-of-the-art performance on the CholecT45 and Cholect50 datasets.
\end{itemize}




\begin{figure*}[t]
\includegraphics[width=\textwidth]{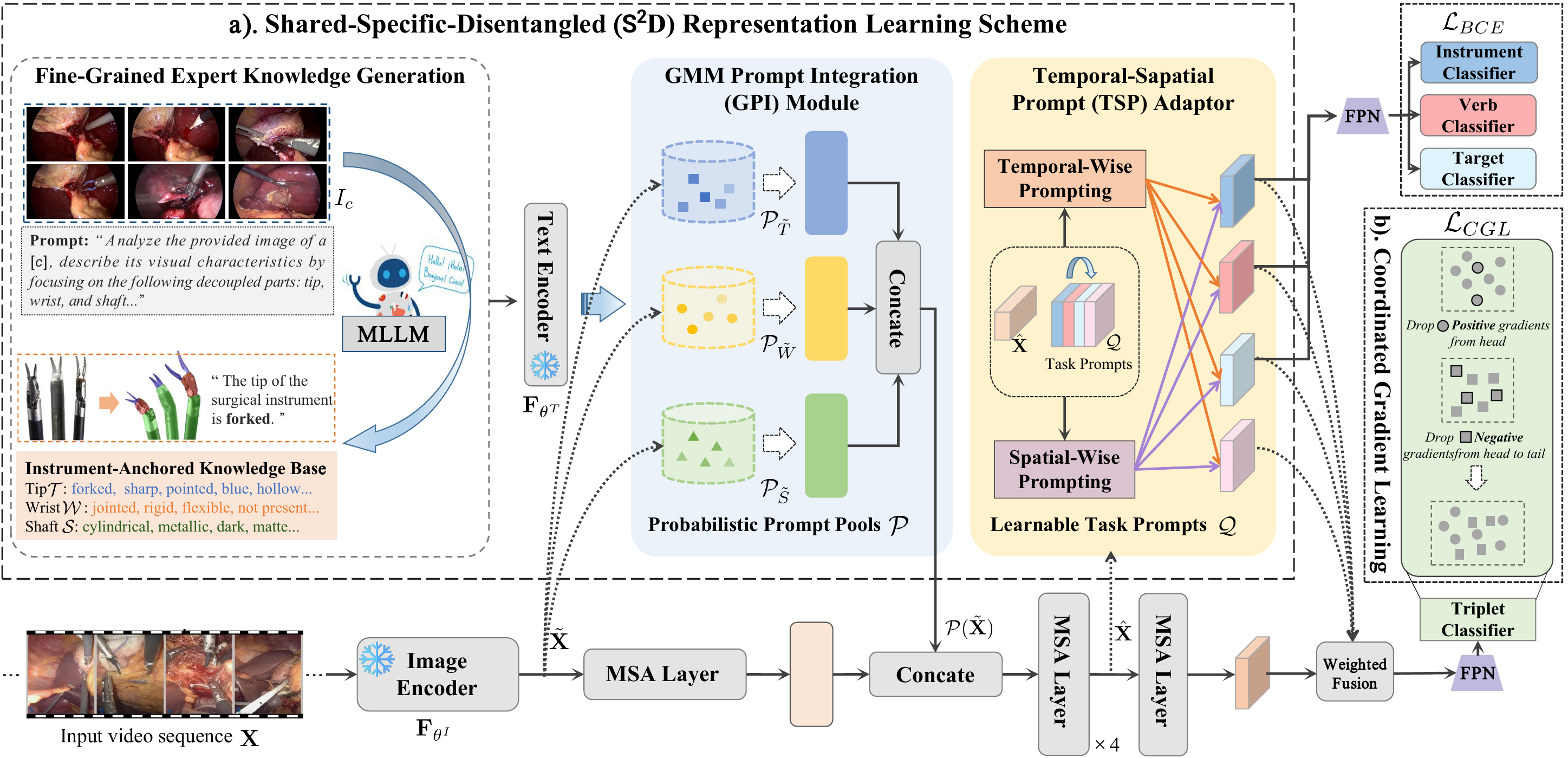}
\caption{Overview of the MEJO framework. Given the input training video sequence $\mathbf{X}$, the GPI module is implemented to enhance visual features with dynamically selected textual semantic prompts from the shared prompt pools $\mathcal{P}$, which are built with fine-grained knowledge generated via MLLM. Next, the TSP adapter is adopted to extract unique features for each task to alleviate inter-task optimization conflicts. To mitigate intra-task conflicts originating from category-level, we propose the CGL strategy by decomposing and rebalancing positive-negative gradients from head and tail classes for the main triplet task.}
\label{img:framework}
\end{figure*}

\section{Related Work}
\label{rw}

\subsection{Surgical Triplet Recognition}
STR is inherently instrument-centric \cite{nwoye2020recognition}, as instruments determine the verbs (actions) and targets (affected structures). A key technique in this paradigm is the use of CAM to generate weakly-supervised localization cues for the identified instruments. For instance, RDV \cite{nwoye2022rendezvous} utilized a class activation guided attention mechanism (CAGAM) to model the semantic relationships between the triplet components. \cite{chen2023surgical} adopted CAM and extended CAGAM in a 3D-CNN network by utilizing the input of 3D tensor data. However, they lack an effective mechanism for integrating expert domain knowledge to enhance the multi-task feature modeling. 
Another stream of methods has targeted the class imbalance issue. MT4MTL-KD \cite{gui2024mt4mtl} employed teacher models trained on less imbalanced sub-tasks (e.g., instrument recognition) to guide a student model tackling the more intricate and imbalanced triplet recognition task. TERL \cite{gui2024tail} utilized a memory bank to store instances for conducting contrastive learning tailored to tail classes. Nevertheless, existing methods overlook the optimization conflicts stemming from multiple tasks, ultimately resulting in suboptimal performance.

\subsection{Multi-Task Learning}
The setting of Multi-Task Learning (MTL) \cite{zhang2021survey} is to learn multiple tasks together through flows of knowledge sharing among tasks. A primary challenge in MTL is the negative transfer, also referred to as inter-task optimization conflicts \cite{sener2018multi, ban2024fair}. Recently, prompt learning \cite{shen2024multitask, ye2022taskprompter, liu2023hierarchical} has been introduced as a solution to leverage the efficiency of prompt tuning within an MTL framework. Despite the potential for knowledge sharing inherent in these approaches, the persistent issue of negative transfer remains unaddressed. In the context of continual learning where tasks come sequentially, \cite{wang2022dualprompt} proposed to learn task-invariant and task-specific objectives with complementary prompts to avoid catastrophic forgetting. While existing methods \cite{cendra2024promptccd, luo2025llm} showcase promising capabilities in the realm of continual learning, their adaptation within the domains of MTL or STR has yet to be explored.

\section{Methodology}

In this section, we present a detailed introduction to the proposed MEJO framework. The overview of MEJO is illustrated in Fig. \ref{img:framework}. First, identifying that severe inter-task optimization conflicts exist in current methods, we introduce the S$^2$D representation learning scheme, aimed at enhancing multi-task feature learning by respectively modeling task-shared and task-specific features. We begin by leveraging MLLM to build an instrument-anchored fine-grained knowledge base $\{\mathcal{T, W, S}\}$. The disentangled knowledge sets are then encoded with a frozen text encoder $\mathbf{F}_{\theta^T}$, represented as $\{\mathcal{\tilde{T}, \tilde{W}, \tilde{S}}\}$. Based on the constructed knowledge base, we build knowledge-driven shared prompt pools $\mathcal{P}$ = $\{\mathcal{P_{\tilde{T}}, P_{\tilde{W}}, P_{\tilde{S}}}\}$ by modeling decoupled knowledge as Gaussian mixture models. Given the input training video sequence $\mathbf{X}$, each frame-wise feature $\tilde{\mathbf{x}}_n$ will be used to retrieve the top-$k$ most relevant prompts ${\mathcal{P}(\tilde{\mathbf{x}}_n)}$ from the shared prompt pools $\mathcal{P}$ through the GMM Prompt Integration (GPI) module. Additionally, the Temporal-Spatial Prompt (TSP) adapter is adopted to extract task-specific features via interaction between visual embeddings $\hat{\mathbf{X}}$ and learnable task-specific prompts $\mathcal{Q}$. The resulting task-specific features will be classified by four distinct learnable classifiers. Specifically, we adopt CGL for the main triplet task that suffers from the most severe long-tailed problem.

\subsection{S$^2$D for Inter-Task Conflicts}
The S$^2$D representation learning scheme is proposed to resolve inter-task optimization conflicts while integrating multi-model knowledge sources. Specifically, it disentangles feature learning into two stages: 1) Learn the informative task-shared representation that is beneficial for all tasks. Probabilistic prompt pooling is adopted to dynamically integrate visual features with retrieved finer-grained instrument-anchored knowledge generated from MLLM. 2) Model individual features for each task to resolve ambiguities. Specifically, we generate unique features for each sub-task considering temporal- and spatial-wise feature interaction.
\subsubsection{Fine-Grained Knowledge Generation} 
Given that STR is instrument-centric, we start by generating instrument-anchored knowledge to enhance the task-shared representations. To build a comprehensive semantic foundation of fine-grained expert knowledge, we employ a systematic prompting strategy to query the MLLM (e.g., GPT-4o). For each surgical instrument class $c$ from $\mathcal{D}$ in a dataset, we present the MLLM with a set of representative images $I_c$ of that instrument for more accurate responses. These images are accompanied by a crafted text prompt designed to elicit disentangled descriptions of the instrument's key functional parts, including candidate attributes of tip $\{t_c\}$, wrist $\{w_c\}$, and shaft $\{s_c\}$. A response template from MLLM is set as:



\tcbset{
    jsonstyle/.style={
        colback = white, colframe = p3,
        fonttitle=\bfseries,
        fontupper=\ttfamily,
        title={Response Template}
    }
}

\begin{tcolorbox}[jsonstyle]
\{
    \hspace*{0mm}"Instrument": [$c$],\\
    \hspace*{4mm}"Attribute": \{\\
    \hspace*{8mm}"tip": [$t_c^1$, $t_c^2$, $t_c^3$, \dots],\\
    \hspace*{8mm}"wrist": [$w_c^1$, $w_c^2$, $w_c^3$, \dots],\\
    \hspace*{8mm}"shaft": [$s_c^1$, $s_c^2$, $s_c^3$, \dots]\\
    \hspace*{4mm}\}\\
\}
\end{tcolorbox}

This extraction process is repeated for all instrument classes in $\mathcal{D}$, resulting in a comprehensive repository of three decoupled knowledge sets $\mathcal{\{T, W, S\}}$, where $\mathcal{T} = \bigcup_{c \in \mathcal{D}} \{t_c\}$, $\mathcal{W} = \bigcup_{c \in \mathcal{D}} \{w_c\}$ and $\mathcal{S} = \bigcup_{c \in \mathcal{D}} \{s_c\}$. A surgeon later verifies the acquired knowledge to ensure their accuracy. Taking the characteristic description ``forked" in the knowledge set $\mathcal{T}$ for example, we then adopt predefined templates such as ``The tip of the surgical instrument is \{forked\}”. This can improve the compatibility of decoupled knowledge with the text encoder (e.g., CLIP \cite{radford2021learning}), ensuring better alignment between textual and visual representations. We also consider the case where there is a background class while no instrument occurs in the image. With ``not present" as descriptions added in both tip, wrist, and shaft knowledge sets, we can implicitly enhance the feature with additional information about the presence of an instrument. Finally, the formulated textual sentences from the knowledge base are converted into high-dimensional vectors $\{\mathcal{\tilde{T}, \tilde{W}, \tilde{S}}\}$ by a pre-trained text encoder $\mathbf{F}_{\theta^T}$.

\subsubsection{GMM Prompt Integration Module} Current prompt tuning methodologies \cite{zhou2022learning, cendra2024promptccd} depend on prompts that are either deterministically or randomly initialized, hindering the seamless integration of expert knowledge. The GPI module introduced aims to enhance the generic task representation by dynamically integrating the most relevant expert knowledge for any given input frame $\mathbf{x}_n$. This process creates a semantic prompt pool containing vector representations for each attribute. As shown in Fig. \ref{img:framework}, prompts are organized into distinct sub-pools $\{\mathcal{P_{\tilde{T}}, P_{\tilde{W}}, P_{\tilde{S}}}\}$ corresponding to the encoded feature vectors $\{\mathcal{\tilde{T}, \tilde{W}, \tilde{S}}\}$. We model the distribution of prompts within each sub-pool using a GMM. A GMM is a probabilistic model that represents a complex distribution as a weighted sum of simpler Gaussian distributions, which has been utilized for prompt pool construction. In this work, each constructed GMM can be interpreted as a comprehensive representation of expert knowledge, offering adaptable insights into image characteristics. This allows us to capture the inherent diversity of different descriptions, thereby augmenting images with flexible and comprehensive descriptions. Considering the sub-pool $\{\mathcal{P_{\tilde{T}}}\}$ with $J$ prompts, the probability density of a feature vector $\mathbf{x}_n$ is given by:
\begin{equation}
    p(\mathbf{x}_n)=\sum_{j=1}^{J}\pi_{j}\mathcal{N}(\mathbf{x}_n|\mu_{j},\Sigma_{j}),
\end{equation}
here, $ \pi_{j} $, $ \mu_{j} $, and $ \Sigma_{j} $ are the mixing coefficient, mean vector, and covariance matrix of the $ j$-th Gaussian component, respectively, each corresponding to one of the $ J $ semantic prompts in the sub-pool. Specifically, $ \pi_{j} $ is set as an average coefficient equal to $1 \over J$. The $ \mu_{j} $ is initialized as its corresponding textual representation, i.e., $ \mu_{j} $ = $\mathcal{\tilde{T}}_j$ for the $ j $-th attribute in $\{\mathcal{\tilde{T}}\}$. The $ \Sigma_{j} $ is implicitly modeled based on the relationship between its textual representation and training images related with $ j $-th attribute denoted as $\mathcal{T}_j$. Specifically, we measure the relationship between $ \mu_{j} $ and images $\mathbf{x} \in \mathcal{T}_j$ using their Euclidean distance.

For an incoming image embedding $\tilde{\mathbf{x}}_n$ from a frozen visual encoder $\mathbf{F}_{\theta^I}$, we find the top $k$ most relevant prompts ${\mathcal{P}(\tilde{\mathbf{x}}_n)} = \{{\mathcal{P}_{\mathcal{\tilde{T}}}(\tilde{\mathbf{x}}_n)}, {\mathcal{P}_{\mathcal{\tilde{W}}}(\tilde{\mathbf{x}}_n)}, {\mathcal{P}_{\mathcal{\tilde{S}}}(\tilde{\mathbf{x}}_n)}\}$ with the highest likelihood values from each decoupled pool, where ${\mathcal{P}_{\mathcal{\tilde{T}}}(\tilde{\mathbf{x}}_n)}$ is denoted as:
\begin{equation}
    {\mathcal{P}_{\mathcal{\tilde{T}}}(\tilde{\mathbf{x}}_n)} = \mu_{\text{top-}k} = \left\{ \mu_j \mid j \in p^{k}
    (\tilde{\mathbf{x}}_n) \right\},
\end{equation}
$p^k$ is a set of the top-$k$ component(s) $j$ from the $ J $ prompts. These selected prompts are used to augment the input sample $\mathbf{x}_n$ as $[\tilde{\mathbf{x}}_n;{P(\tilde{\mathbf{x}}_n)}]$. We then apply prefix-tuning \cite{li2021prefix} to get the enhanced representations $\hat{\mathbf{X}} \in \mathbb{R}^{L \times E} $, where $L$ represents the sequence length.


\subsubsection{Temporal-Spatial Prompt Adapter} 
\begin{figure}[t]
\centering
\includegraphics[width=8.5cm]{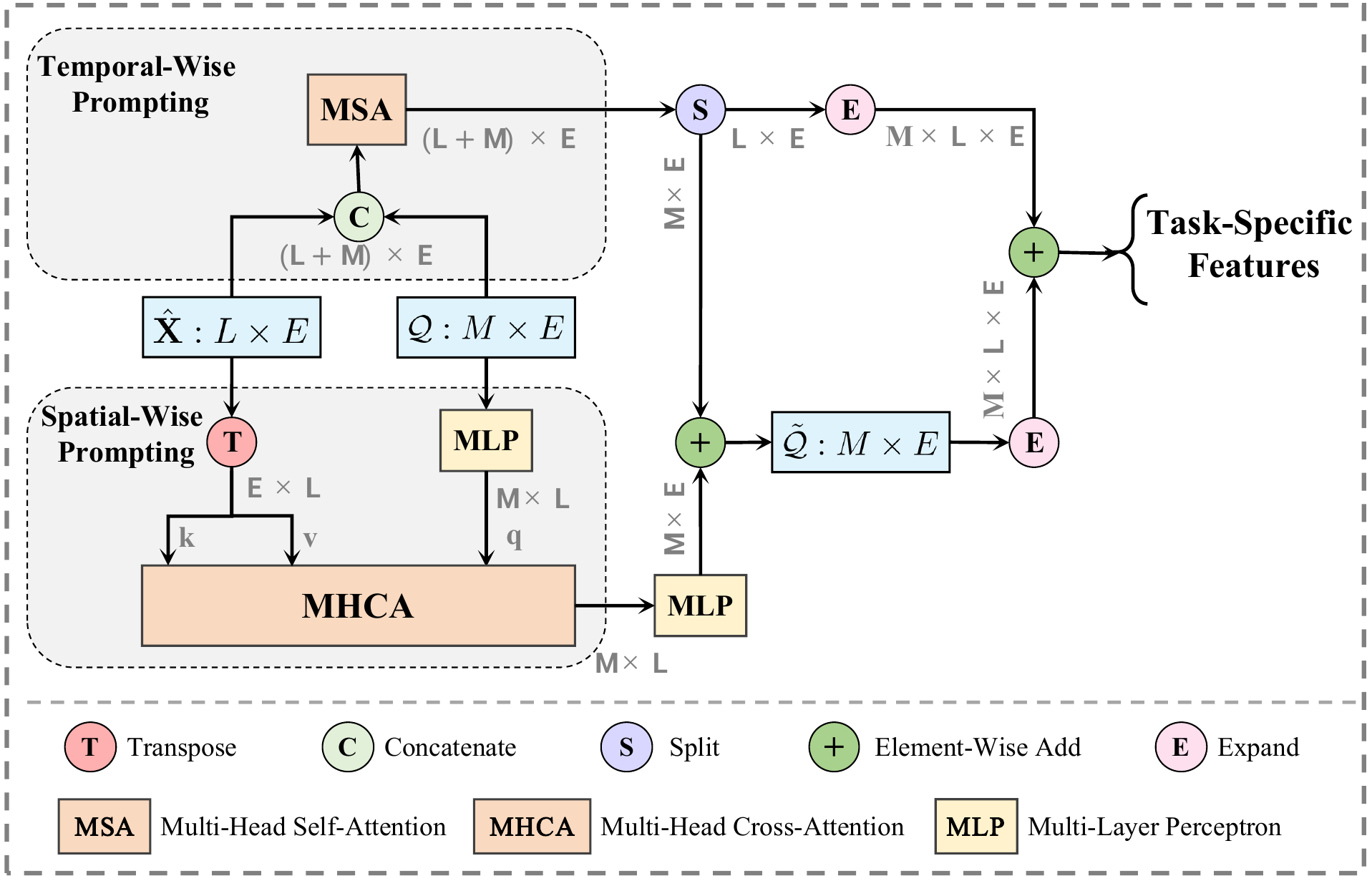}
\caption{An illustration of the Temporal-Spatial Prompt Adapter. The TSP adapter learns the temporal- and spatial-wise task-specific features with no need for duplicate network designs for each task.}
\label{img:stp}
\end{figure}

Previous methods, such as TERL \cite{gui2024tail}, built task-specific branches with distinct classifiers without modeling specific features for each task, resulting in inter-task optimization competition. To tackle this issue, we extract unique features for each task through the TSP adapter. As illustrated in Fig. \ref{img:stp}, distinct task prompts are embedded with the task-generic video features to operate attention mechanisms from both spatial and temporal dimensions for more comprehensive feature interaction. We randomly initialize a small set of dedicated, learnable prompt tokens $\mathcal{Q} \in \mathbb{R}^{M \times E}$, where $M$ is the number of tasks and $E$ is the dimension of each prompt embedding. Given the video-wise feature $\hat{\mathbf{X}} \in \mathbb{R}^{L \times E}$, we first conduct temporal-wise prompting. Specifically, we concatenate the feature map and prompt tokens and feed them into a multi-head self-attention (MSA) layer:
\begin{equation}
\mathbf{F}_{t},\mathbf{P}_{t}=\operatorname{Split}(\operatorname{MSA}(\operatorname{Concat}(\hat{\mathbf{X}},\mathcal{Q}))),
\end{equation}
where $\mathbf{F}_{t}\in\mathbb{R}^{L\times E}$ is the basic temporally-enhanced feature map and $\mathbf{P}_{t}\in\mathbb{R}^{M\times E}$ represents the temporally-enhanced prompt tokens. Concurrently, to capture spatial information, we employ a multi-head cross-attention (MHCA) mechanism. The learnable prompts $\mathcal{Q}$ are transformed by a multilayer perceptron (MLP) to form the query $q \in \mathbb{R}^{M \times L}$, while the transposed video-wise feature $\hat{\mathbf{X}}^\top \in \mathbb{R}^{E \times L}$ serves as the key $k$ and value $v$. The resulting spatially aware prompts are computed by:
\begin{equation}
\mathbf{P}_{s}=\operatorname{MLP}(\operatorname{MHCA}(\operatorname{MLP}(\mathcal{Q}),\hat{\mathbf{X}}^\top,\hat{\mathbf{X}}^\top)),
\end{equation}
where $\mathbf{P}_{s}\in\mathbb{R}^{M\times E}$. Finally, the temporal and spatial prompts are integrated through element-wise addition to form the final task-specific prompts $\tilde{\mathcal{Q}}=\mathbf{P}_{t}+\mathbf{P}_{s}$.
To generate the task-specific features, we expand both the base feature map and the refined prompts and combine them:
\begin{equation}
\mathbf{Z}=\operatorname{Expand}(\mathbf{F}_t)+\operatorname{Expand}(\tilde{\mathcal{Q}})\in\mathbb{R}^{M\times L\times E},
\end{equation}
where $\mathbf{Z} = \{\bm{z_i}, \bm{z_v}, \bm{z_t}, \bm{z_{ivt}}\} $ and $M = 4$. 
The final triplet feature is weighted fused as:
\begin{equation}
\bm{z_{ivt}}= \alpha(\bm{z_i}+ \bm{z_v}+ \bm{z_t}+ \bm{z_{ivt}})+\operatorname{MSA}(\hat{\mathbf{X}}),
\end{equation}
where $\alpha$ is a hyperparameter for weighted summation. The generated task-specific features are then fed to task-specific classifiers for loss calculation.

\subsection{CGL for Intra-Task Optimization Conflicts}
Due to the severe long-tailed data distribution in CholecT45, conventional binary cross-entropy (BCE) loss performs poorly in STR. To better investigate the detailed optimization procedure, we dissect the vanilla BCE loss into positive $\mathcal{L}^+$ and negative $\mathcal{L}^-$ components \cite{tan2020equalization}. Let $\mathcal{O} = \{(\mathbf{x}_n, {y}_n)\}_{n=1}^\mathcal{N}$ be a batch of data where $\mathbf{x}_n$ is an input instance and ${y}_n \in \{0,1\}^\mathcal{G}$ is its corresponding label vector for $\mathcal{G}$ categories. The decomposed loss can be denoted as: 
\begin{equation}
\begin{aligned}  
\mathcal{L}_{\text{BCE}} = -\sum_{n=1}^{\mathcal{N}} \sum_{g=1}^\mathcal{G} \underbrace{y_{n,g}\log\sigma(\bm{w}_g^\top\bm{z_{ivt}}^n)}_{\mathcal{L}^+} + \\
\underbrace{(1-y_{n,g})\log(1-\sigma(\bm{w}_g^\top\bm{z_{ivt}}^n))}_{\mathcal{L}^-},
\end{aligned}
\end{equation}
where $\bm{\omega}_g \in \mathbb{R}^E$ represents the classifier weights for triplet category $g$ and $\sigma(\cdot)$ is the sigmoid function. For clarity, we omit the instance index $n$ and task index $ivt$, and focus on a single triplet class $g$. Suppose the logit $z_g = {w}_g^\top\bm{z_{ivt}}^n$, the gradient of BCE loss with respect to $z_g$ is:
\begin{equation}
\frac{\partial \mathcal{L}_{BCE}}{\partial z_g} = \sigma(z_g) - y_g.
\end{equation}
For a positive sample ($y_g = 1$), the gradient is $|\sigma(z_g) - 1|$, and a negative sample ($y_g = 0$), the gradient is $\sigma(z_g)$. As illustrated in Fig. \ref{img:grad}, the gradient gap between negative and positive losses in tail classes are significant. For a more balanced training gradient, such as in head classes, a simple but effective solution is to speed up positive label learning and slow down negative label learning in tail classes. We achieve this by balancing the ratio of positive-to-negative gradients. 

Instead of simply discarding the negative loss of tail classes to slow down the optimization of negative labels as in EQ Loss \cite{tan2020equalization}, we propose two sophisticated improvements based on the experimental analysis on BCE loss discussed in Section \ref{sec:fgl}. 1) $h_g^-$: Suppress the negative loss solely from head classes for tail classes by disregarding gradients (with probability $\gamma$) from head category samples for tail categories.  2) $h_g^+$: Discard gradients (with probability $\gamma$) of the positive loss from head classes. Slowing down positive learning in head classes offers an alternative approach to accelerate learning in tail classes, given the competitive nature of these two loss components. The CGL loss function for one input instance $\mathbf{x}_n$ is denoted as follows:
\begin{equation}
    \mathcal{L}_{\text{CGL}} = -\sum_{g=1}^{\mathcal{G}} h_{g}^+y_g \log\left(\sigma({z}_{g})\right) + h_{g}^-(1-y_g) \log\left(1-\sigma({z}_{g})\right),
\end{equation}
where ${h}_g^+ = 1-\lambda \mathcal{E}(g)$ and ${h}_g^- = 1 - \lambda \mathcal{F}(g)$. $\mathcal{E}(g)$ is 1 if $g$ belongs to head category and 0 otherwise. $\mathcal{F}(g)$ is 1 if $g$ belongs to tail category and $x_n$ belongs to head class, otherwise is 0. 
$\lambda$ is a random variable with a probability of $\gamma$ to be 1 and $1-$ $\gamma$ to be 0.

\begin{table*}[t]
\centering
\fontsize{10pt}{10pt}\selectfont
    \renewcommand{\arraystretch}{1.2}
    \setlength{\tabcolsep}{2mm}{
\begin{tabular}{@{}lccccccc@{}}
\toprule
Method & Backbone & $AP_I$ & $AP_V$ & $AP_T$ & $AP_{IV}$ & $AP_{IT}$ & $AP_{IVT}$ \\
\midrule
TripNet \cite{nwoye2020recognition} & ResNet-18 & 89.9±1.0 & 59.9±0.9 & 37.4±1.5 & - & - & 24.4±4.7 \\

RDV \cite{nwoye2022rendezvous} & ResNet-18 & 89.3±2.1 & 62.0±1.3 & 40.0±1.4 & 34.0±3.3 & 30.8±2.1 & 29.4±2.8 \\
RiT \cite{sharma2023rendezvous} & ResNet-18 & 88.6±2.6 & 64.0±2.5 & 43.4±1.4 & 38.3±3.5 & 36.9±1.0 & 29.7±2.6 \\
TripDis \cite{chen2023surgical} & ResNet-50 & 91.2±1.9 & 65.3±2.8 & 43.7±1.6 & - & - & 33.8±2.5 \\
SelfD \cite{yamlahi2023self} & SwinB×2+SwinL & - & - & - & - & - & 38.5±0.0 \\
MT4MTL-KD \cite{gui2024mt4mtl} & ResNet-18+SwinL & \underline{93.9±2.0} & \textbf{73.8±2.0} & \textbf{52.1±5.2} & 46.5±3.4 & 46.2±2.3 & 38.9±1.6 \\

TERL-T \cite{gui2024tail} & SwinT+MSTCN & 93.1±2.4 & 71.1±1.7 & 48.9±3.9 & 44.9±4.4 & 41.9±3.1 & 35.7±2.3 \\
TERL-B \cite{gui2024tail} & SwinB+MSTCN & 93.5±2.4 & 72.8±2.8 & 51.3±3.8 & \underline{47.0±5.6} & \underline{45.7±2.8} & 38.9±2.5 \\
\midrule
Focal Loss \cite{lin2017focal} & SwinT+TransFPN & 92.6±2.2 & 69.8±1.4 & 47.3±5.4 & 45.7±4.9 & 43.4±2.9 & 38.0±2.9 \\
CB Loss \cite{cui2019class} & SwinT+TransFPN & 93.1±2.8 & 67.5±5.3 & 48.0±5.3 & 45.1±4.8 & 43.6±2.2 & 37.7±2.0 \\
EQ Loss \cite{tan2020equalization} & SwinT+TransFPN & 92.5±2.9 & 69.4±1.7& 47.8±3.8 &  45.2±5.3 & 43.4±1.6 & 37.8±3.2 \\
EQ Loss v2 \cite{tan2021equalization} & SwinT+TransFPN & 89.5±2.7 & 65.8±2.6 & 41.7±1.9 & 40.8±3.6 & 40.6±2.8 & 37.3±2.3 \\
\midrule
\textbf{MEJO-T (Ours)} & SwinT+TransFPN & 93.4±2.0 & 70.9±1.5 & 49.9±5.3 & 46.3±5.3 & 45.3±1.9 & \underline{40.1±3.6} \\
\textbf{MEJO-B (Ours)} & SwinB+TransFPN & \textbf{93.9±2.2} & \underline{72.9±1.8}  & \underline{51.6±5.7} & \textbf{47.8±6.4} & \textbf{46.3±1.8} & \textbf{41.2±2.6}  \\
\bottomrule
\end{tabular}
}
\caption{Quantitative comparisons between the proposed method and the SOTA methods using 5-fold cross-validation on the CholecT45 dataset. The mean and standard deviation results of AP are reported. \textbf{Best} and \underline{Second Best}.}
\label{tab:benchmark}
\end{table*}

\section{Experiment}\label{sec:exp}
\subsection{Datasets and Implementation Details}
We conduct experiments on two public datasets from the CholecTriplet2021 challenge \cite{nwoye2023cholectriplet2021}. The CholecT45 dataset \cite{nwoye2022data} contains 45 laparoscopic cholecystectomy video sequences comprising 100.9K frames annotated with 161K triplet instance labels. Each frame includes annotations of 100 binary action triplets, consisting of 6 instruments (${I}$), 10 verbs (${V}$), and 15 targets (${T}$). We adopt the official 5-fold cross-validation strategy with a 31-5-9 split for training, validation, and testing, respectively. This paper follows \cite{gui2024tail} to perform ablation studies and sensitivity analysis on Fold 1. We also validate our method using the CholecT50 dataset, an extension of the CholecT45 dataset with five additional videos. Consistent with the data partitioning scheme of \cite{xi2023chain}, we allocate 40 videos for training, 5 videos for validation, and 5 videos for testing. Performance is evaluated using average precision (AP) metrics including triplet AP ($AP_{IVT}$), association AP (${AP}_{IV}$ and ${AP}_{IT}$), and component AP (${AP}_{I}$, ${AP}_{V}$, ${AP}_{T}$), where ${AP}_{IVT}$ serves as the primary metric for complete triplet recognition.

We first adopt a pretrained Swin transformer as the spatial visual encoder \cite{gui2024tail} to extract image features, then construct a temporal transformer backbone termed as TransFPN that combines design elements from Actionformer~\cite{zhang2022actionformer} and FPN~\cite{lin2017feature}. TransFPN consists of 6 base multi-head self-attention (MSA) layers followed by a Feature Pyramid Network (FPN) module with 5 additional MSA layers (scale factor~$=2$). The model processes complete video-wise feature embeddings ($E=768$ for SwinT or $E=1024$ for SwinB) from the frozen visual encoder. The class categorization is based on sample counts: head classes ($>\!10,\!000$ samples), tail classes ($<\!1,\!000$ samples), yielding a head-medium-tail distribution ratio of 3:13:84. We employ pre-trained CLIP as the text encoder. All models are trained for 800 epochs using SGD with momentum $\mu=0.95$ and initial learning rate $5\!\times\!10^{-2}$, implemented on a single NVIDIA RTX 3090 GPU for both training and inference. We select $\lambda = 0.1$ and $\alpha = 0.1$ as they emerge as optimal choices based on the sensitivity analysis provided in the Appendix.
\subsection{Experimental Results}
\subsubsection{Main Results}
In the Cholect45 dataset, we compare the proposed framework with 8 state-of-the-art (SOTA) triplet recognition methods and 4 representative cost-sensitive methods designed for training with imbalanced data. As shown in Table \ref{tab:benchmark}, the experimental results highlight the superior performance of our method compared to the competing approaches. In our comparison, MEJO-T and MEJO-B denote the utilization of SwinT or SwinB as the pretrained visual encoder, respectively. Specifically, MEJO-T achieves an average $AP_{IVT}$ of 40.1\%, surpassing TERL-T by 4.4\%. MEJO-B achieves the highest performance at 41.2\%, outperforming the second-best method (TERL-B) by 2.3\%. Moreover, our method excels in instrument recognition ($AP_{I}$), instrument-verb association recognition ($AP_{IV}$), and the instrument-target association recognition ($AP_{IV}$), demonstrating a clear advantage in tasks related to instruments and supporting the effectiveness of our instrument-anchored knowledge integration. Across individual component recognition, our method demonstrates competitive performance, achieving 93.9\%, 72.9\%, and 51.6\% in terms of $AP_{I}$, $AP_{V}$, $AP_{T}$, showcasing the advantage of our framework in managing multiple tasks. Notably, traditional cost-sensitive approaches combined with our proposed backbone achieve approximately 38\%, underscoring their limited utility in STR.

\begin{figure}[t]
\centering
\includegraphics[width=8cm]{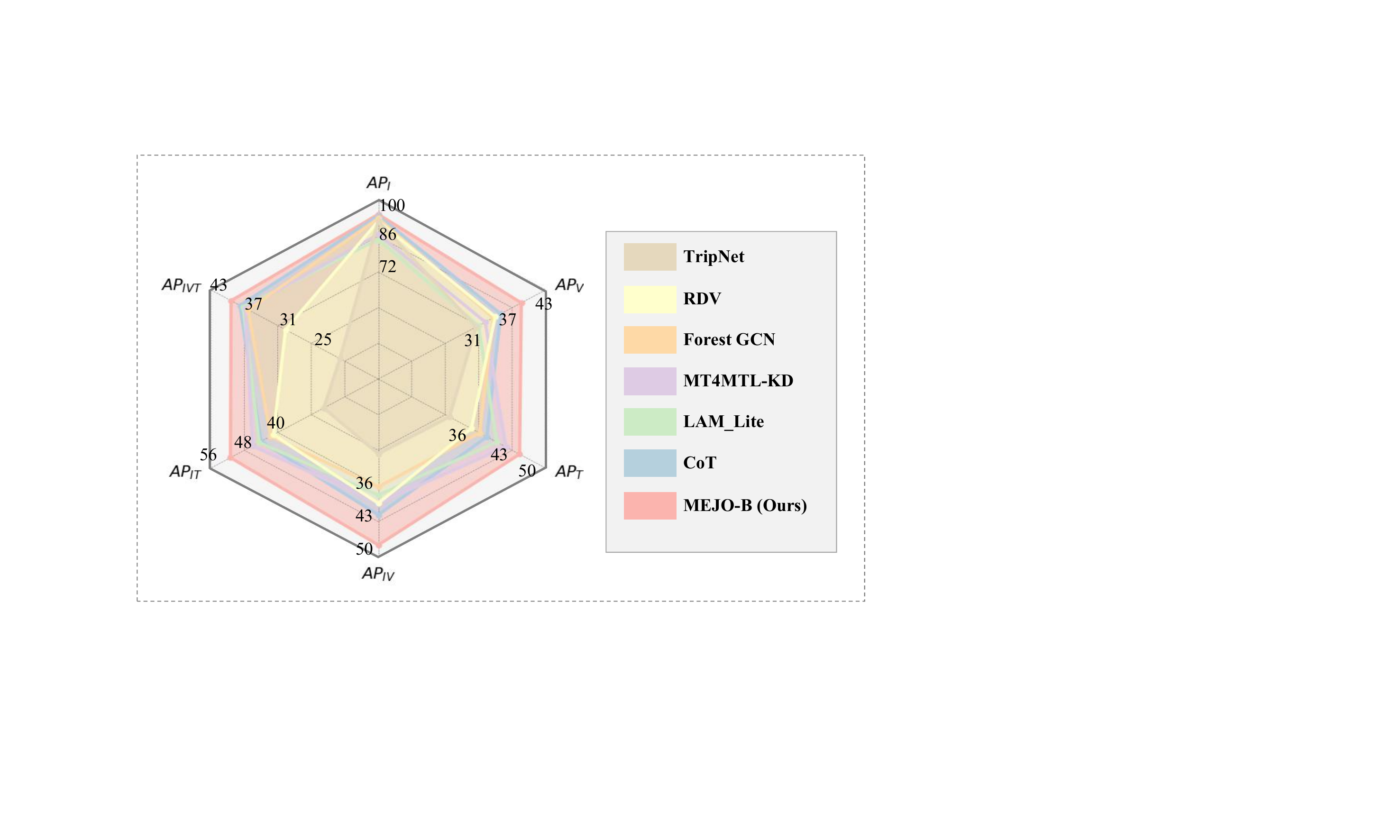}
\caption{Validating our method on the CholecT50 dataset.}
\label{img:t50}
\end{figure}

For the evaluation on the Cholect50 dataset, we employ radar charts to visually compare the performance of our approach against 7 SOTA methods, with each method represented by different colored fill areas. The compared methods include Forest GCN \cite{xi2022forest}, CoT \cite{xi2023chain}, TCN \cite{nwoye2022rendezvous}, LAM\_Lite \cite{li2024parameter}, etc. As illustrated in Fig. \ref{img:t50}, the proposed MEJO-B outperforms the second-best method (CoT) by 1.6\% in $AP_{IVT}$ and consistently achieves the best performance across 6 metrics, demonstrating its effectiveness in addressing the STR challenge.

\subsection{Ablation Study}
\subsubsection{Ablation on Different Modules}The results of the ablation study in Table \ref{tab:ablation} demonstrate the consistent improvements achieved by using the TransFPN backbone, as well as progressively incorporating three novel modules. Following TERL \cite{gui2024tail}, we adopt a four-stage TCN (MSTCN) as a baseline method with $AP_{IVT}$ equal to $37.35\%$. The TransFPN backbone exhibits a better capability over the baseline model with a +0.9\% increase in $AP_{IVT}$ score, although it demonstrates lower performance in $AP_I$ and $AP_V$ metrics. Incorporating the CGL strategy initiates a comprehensive enhancement across all six metrics, underscoring its pivotal role in mitigating intra-task conflicts within STR. Notably, the integration of the GPI module achieves an impressive $AP_{IVT}$ score of 41.03\%. To validate the impact of our designed knowledge base, an evaluation of the GPI module with randomly initialized prompts reveals a -0.8\% decrease in $AP_{IVT}$, emphasizing the critical nature of knowledge integration. The culmination of all components results in the highest performance of 42.25\% with a substantial improvement of +4.9\%, indicating the effectiveness of these integrated modules.

\begin{table}[t]
    \centering
    \fontsize{9pt}{9pt}\selectfont
    \setlength{\tabcolsep}{1mm}{
\begin{tabular}{@{}cccc|cccccc@{}}
\toprule
TFPN & C. & G. & T. & $AP_I$ & $AP_V$ & $AP_T$ & $AP_{IV}$ & $AP_{IT}$ & $AP_{IVT}$ \\
\midrule
{}  & {} & {} &   & 90.3 & 71.3 & 52.2 & 42.9 & 45.0 & 37.4 \\
\midrule
 \checkmark & {} & {}  & & 89.4 & 67.2 & 53.2 & 42.1 & 45.1 & 38.3  ($\uparrow$0.9) \\
 {\checkmark} & \checkmark & {}&   & 90.8& 68.8 & 55.6 & 42.3 & 47.1 &  40.0 ($\uparrow$2.7) \\
 \checkmark & \checkmark &   & \checkmark & 91.2 & 68.8 & 57.0 & 42.5 & \underline{50.9} & 41.0 ($\uparrow$3.6) \\
\checkmark & \checkmark & \checkmark  & & \textbf{91.7} & \underline{70.0} & \underline{58.0} & \textbf{44.0} & 49.6 & \underline{41.1} ($\uparrow$3.7) \\
\checkmark & \checkmark & \checkmark  & \checkmark &  \underline{91.6} & \textbf{71.2} & \textbf{58.1} & \underline{43.8} & \textbf{51.1}  & \textbf{42.3} ($\uparrow$4.9) \\
\bottomrule
\end{tabular}
}
\caption{Ablation study on module contributions of our framework (MEJO-T). The first row represents the results of the baseline method TERL using MSTCN. (TFPN: TransFPN, C.: CGL, G.: GPI, T.: TSP)}
\label{tab:ablation}
\end{table}

\subsubsection{Ablation on CGL} \label{sec:fgl}
\begin{figure}[t]
\centering
\includegraphics[width=8cm]{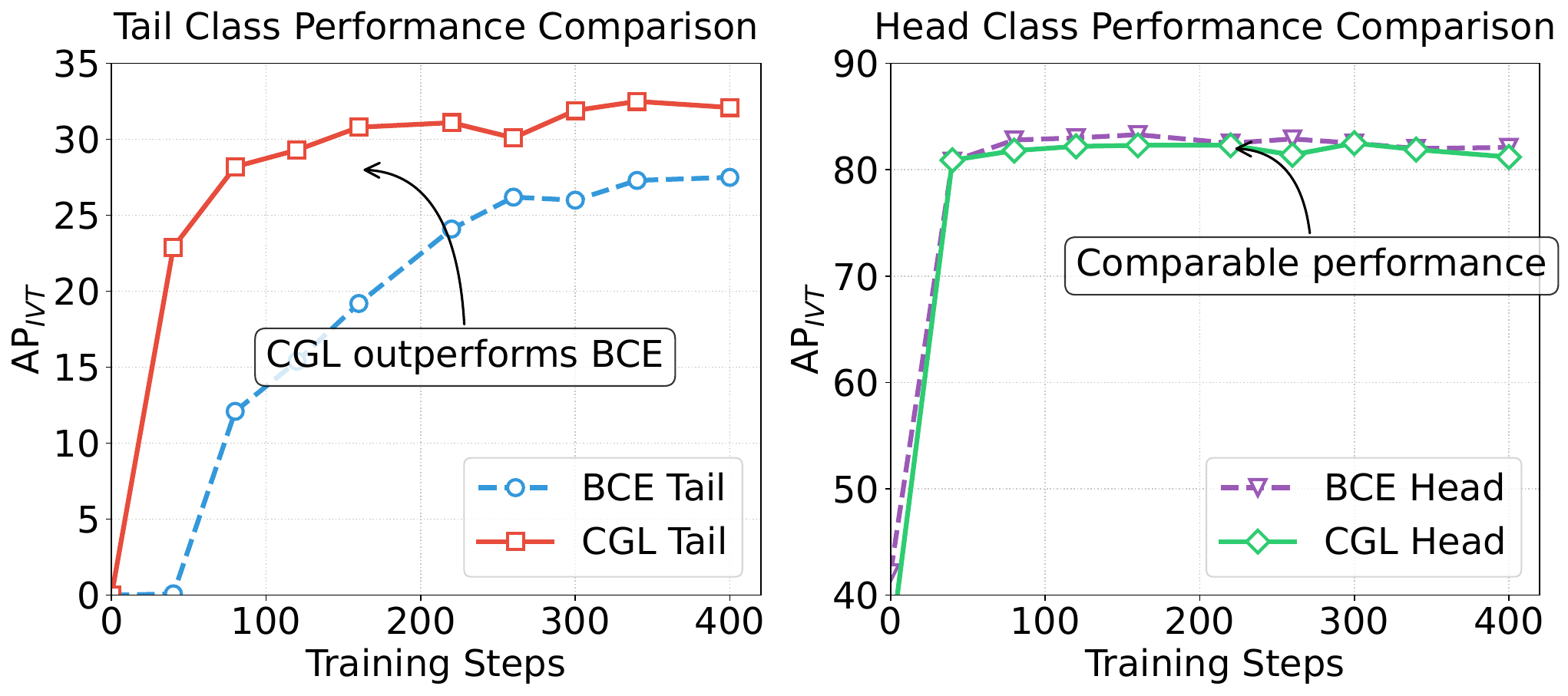}
\caption{Performance comparison between BCE and CGL. }
\label{img:fgl}
\end{figure}

\begin{figure}[t]
\includegraphics[width=8cm]{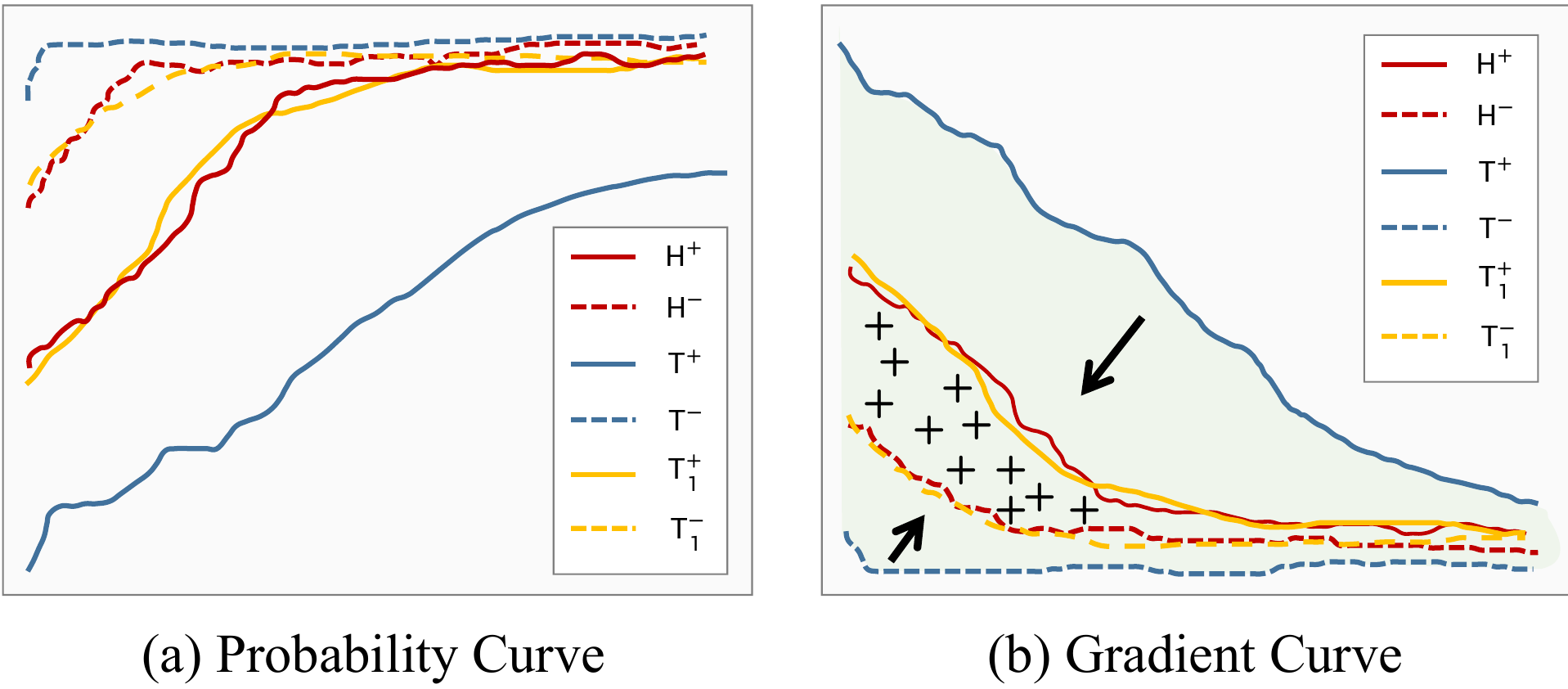}
\caption{Predicted mean probability (a) and mean gradient (b)
curves on the dataset CholecT45 during training. H indicates the head classes, and T denotes the tail classes. The $^+$ and solid line represent the probability and gradient of positive labels, while the $^-$ and dotted line indicate the probability and gradient of negative labels. T$^+_1$ and T$^-_1$ indicate the gradient trained with CGL, and the space filled with \textbf{+} represents the gradient gap between them.}
\label{img:grad}
\end{figure}

To delve into CGL functionality, we showcase mean probability and gradient curves using BCE and CGL in Fig. \ref{img:grad}. With BCE training (Fig. \ref{img:grad}(a)), tail classes exhibit rapid learning in negative probability but slower growth in positive probability, indicating that the tail classifiers are biased towards negative samples and prone to recognizing true positive samples as negative ones. The head classifiers represent a relatively more balanced trend between positive and negative probabilities. In subplot (b), a significant gradient gap (shaded green) highlights the contrast in gradient magnitudes between positive and negative labels for tail classes. This disparity in gradient flow potentially leads to suboptimal optimization. The CGL strategy denoted by T$^+_1$ and T$^-_1$ effectively establishes a more balanced optimization landscape similar to that of head classes, enhancing tail performance while maintaining comparable head performance, as shown in Fig. \ref{img:fgl}. 

\subsubsection{Ablation on Shared-Prompt Pool Location}  \begin{table}[h]
    \centering
    \fontsize{9pt}{9pt}\selectfont
    \setlength{\tabcolsep}{2mm}{
    \begin{tabular}{@{}ccccccc@{}}
        \toprule
        Location & $AP_{I}$ & $AP_{V}$ & $AP_{T}$ & $AP_{IV}$ & $AP_{IT}$ & $AP_{IVT}$ \\
        \midrule
        {0} & 91.44 & 70.94 &  55.27&  44.11 & 47.52 & \underline{40.89} \\
        1 & \underline{91.45} &  \textbf{71.15} & \textbf{56.12}  & \underline{44.44} & 47.66 &  \textbf{40.91} \\
        2 &  91.42& 70.43 & 55.02 &  43.81& \textbf{48.91} & 40.63 \\
        3 & 91.22 & \underline{71.04} & 54.41 &  44.19&  47.70 & 40.51\\
        4 & \textbf{91.63} &  71.02 &  \underline{55.29} & \textbf{44.49} & \underline{48.33} & 40.55\\
        \bottomrule
    \end{tabular}}
    \caption{Ablation on varying shared-prompt pool locations.}
    \label{tab:location}
\end{table}
Table \ref{tab:location} systematically assesses the impact of varying shared-prompt pool locations on the performance of STR. The location number $l \in \{0,1,2,3,4\}$ denotes the index of the MSA layer integrated with the GPI module. The results underscore that at $l=1$, the framework achieves the best performance, yielding an $AP_{IVT}$ score of 40.91\%. Remarkably, the $AP_{IVT}$ metric demonstrates stable performance across different locations, indicating the robustness of the proposed GPI module at various feature levels.



\section{Conclusion}

In conclusion, surgical triplet recognition presents significant challenges due to inter-task optimization conflicts from inadequate representation modeling and intra-task conflicts caused by long-tailed data distributions. We propose the MEJO framework to tackle both challenges simultaneously. For the first challenge, we leverage the S$^2$D learning scheme to respectively model task-shared and task-specific representations. Shared features are dynamically enhanced by a GPI module that integrates fine-grained expert knowledge from MLLM. Furthermore, the TSP adapter captures non-shared features while modeling cross-task interactions. In tackling intra-task conflicts, we introduced the CGL strategy to effectively rebalance gradients between head and tail classes. Future efforts could prioritize cultivating in-context understanding with MLLM to enhance zero-shot application and enable more generalized usage across various scenarios.

\bibliography{aaai2026}

\end{document}